# Autocodificadores Variacionales (VAE) Fundamentos Teóricos y Aplicaciones

Survey

**Jordi de la Torre**[*]
Ph.D. in Computer Science (ML/AI)
Universitat Oberta de Catalunya
Barcelona, ES
jordi.delatorre@gmail.com

February 18, 2023

## Abstract

Los VAEs son modelos gráficos probabilísticos basados en redes neuronales que permiten la codificación de los datos de entrada en un espacio latente formado por distribuciones de probabilidad más sencillas y la reconstrucción, a partir de dichas variables latentes, de los datos de origen. Después del entrenamiento, la red de reconstrucción, denominada decodificadora, es capaz de generar nuevos elementos pertenecientes a una distribución próxima, idealmente igual, a la de origen. Este artículo se ha redactado en español para facilitar la llegada de este conocimiento científico a la comunidad hispanohablante.

## Abstract

VAEs are probabilistic graphical models based on neural networks that allow the coding of input data in a latent space formed by simpler probability distributions and the reconstruction, based on such latent variables, of the source data. After training, the reconstruction network, called decoder, is capable of generating new elements belonging to a close distribution, ideally equal to the original one. This article has been written in Spanish to facilitate the arrival of this scientific knowledge to the Spanish-speaking community.

---

[*]mailto:jordi.delatorre@gmail.com web:jorditg.github.io

# 1 Autocodificadores

Antes de presentar los VAEs, consideramos apropiado brindar una breve introducción a la estructura general de los auto-codificadores. Esta arquitectura precede en la historia a los variacionales y comparte muchos aspectos en común, aunque sus aplicaciones no se enfoquen en la generación de datos como sucede con los VAEs.

## 1.1 Introducción

Los auto-codificadores [1], [2], [3] son un tipo de red neuronal especializada en la representación de un espacio dimensional de origen en otro más pequeño. El objetivo de esta transformación es la representación de la información observada en un vector de menor dimensión que lo represente con la mayor fidelidad posible. Se pretende que dicha compresión sirva para codificar en la red una representación significativa de los verdaderos factores explicativos de las señales observadas. A la representación vectorial reducida se le denomina espacio latente.

Los auto-codificadores (fig. 1.1) son una propuesta de arquitectura para llevar a cabo este proceso. Estos consisten en dos partes diferenciadas: un codificador que transforma el vector de datos inicial en una representación interna y un decodificador que a partir de esta última intenta reconstruir el original. Ambos elementos son redes neuronales la función de las cuales es modelizar la distribución de probabilidad de los datos de origen. La arquitectura se optimiza de forma global. El cuello de botella existente en la zona del espacio latente es el que, idealmente, obliga a la arquitectura a comprimir en representaciones más sencillas la información relevante a partir de la cual se reconstruirá el original.

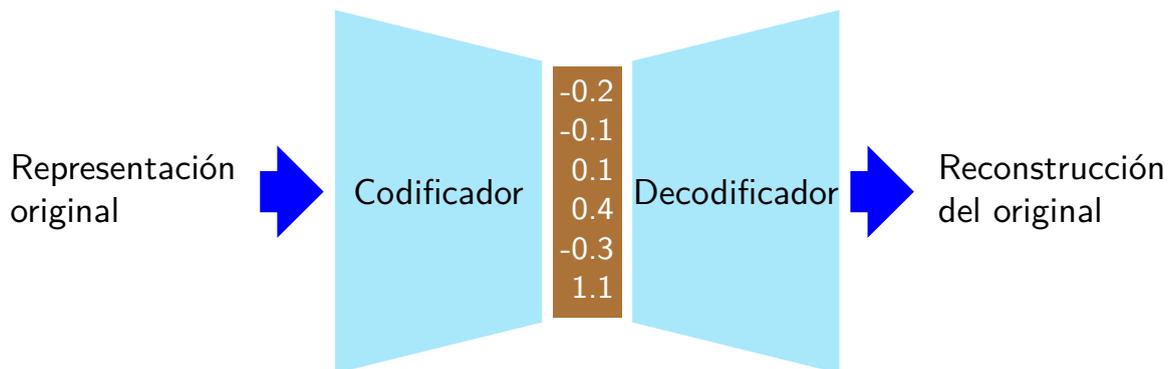

Figura 1: Esquema general de un auto-codificador

## 1.2 Definición formal

Sea $\mathcal{X}$ el espacio de los datos de entrada el $\mathcal{Z}$ el de la representación codificada, ambos espacios euclidianos, esto es $\mathcal{X} = \mathbb{R}^m$, $\mathcal{Z} = \mathbb{R}^n$, donde habitualmente $m > n$.

Definimos dos funciones paramétricas denominadas codificador $E_\phi : \mathcal{X} \to \mathcal{Z}$, parametrizada por $\phi$ y decodificador $D_\theta : \mathcal{Z} \to \mathcal{X}$, parametrizada por $\theta$.

Para cualquier $x \in \mathcal{X}$, denotamos como $z = E_\phi(x)$ a la variable latente $z \in \mathcal{Z}$ derivada de aplicar la función $E\phi$ a $x$ y a $x' = D_\theta(z)$ como la reproducción de $x$ mediante decodificación de $z$. Habitualmente las funciones paramétricas $E_\phi$ y $D_\theta$ son redes neuronales.

## 1.3 Entrenamiento

Para optimizar la arquitectura se debe definir una objeto matemático a optimizar. Para ello definimos inicialmente la función distancia $d$ como $d : \mathcal{X} \times \mathcal{X} \to [0, \infty]$, tal que $d(x, x')$ es una función distancia que mide la diferencia entre ambos vectores.

Definida $d$, definiremos la función de optimización que minimiza la distancia sobre todo el conjunto de datos. Esto se puede hacer minimizando el valor esperado de la distancia sobre todo el conjunto de datos de origen:

$$L(\theta, \phi) = arg \min_{\theta, \phi} \mathbb{E}_{x \sim \mu_{ref}}[d(x, D_\theta(E_\phi(x)))]$$

Para aquellos casos en que disponemos únicamente de un conjunto de datos finito representativo de $\mathcal{X}$, $x_1, ..., x_N \in \mathcal{X}$ entonces $\mu_{ref} = \frac{1}{N} \sum_{i=1}^{N} \delta_{x_i}$.

Si se define $d$ como $d(x, x') = ||x - x'||^2$ entonces el problema se convierte en una optimización de mínimos cuadrados.

### 1.4 Auto-codificadores regularizados

El objetivo último del auto-codificador es conseguir una representación comprimida del vector de entrada, idealmente formada por sus factores constitutivos. Esto se pretende conseguir optimizando un objetivo matemático, aunque no siempre la optimización del segundo conduce al primero. Debido a que la optimización se realiza mediante el uso de una muestra finita de la población total, esto es, con un conjunto de entrenamiento finito, en determinadas circunstancias puede darse el caso que a la función paramétrica le sea más sencilla la memorización del conjunto de datos que el aprendizaje de representaciones sencillas. En estos casos, decimos que el auto-codificador aprende la función identidad. Este caso no es deseable, pues se aleja del objetivo que pretendemos conseguir.

Una forma de evitar esto es dotar a la función de optimización de mecanismos de regularización que eviten la memorización. Dependiendo de cómo se realice esta regularización distinguimos entre distintas subclases de auto-codificadores, entre los que se encuentran los auto-codificadores poco densos, los de atenuación de ruido y los contractivos.

#### 1.4.1 Auto-codificadores poco densos (SAE)

Los auto-codificadores poco-densos introducen una restricción en la codificación de manera que esta sea poco densa, en el sentido que únicamente se permita que unos pocos de los elementos de la representación latente estén activos. Hay dos formas básicas de conseguir este objetivo: la primera es fijar a cero todas las activaciones excepto las $k$ mayores, donde $k$ es un valor predefinido. A este se les denomina auto-codificador poco denso k [4]; la segunda consiste en añadir a la función de pérdida original una función de regularización, esto es, $\min_{\theta, \phi} L(\theta, \phi) + \lambda L_{sparsity}(\theta, \phi)$, donde $\lambda > 0$ fija cuánta baja densidad queremos aplicar [5]. Para un auto-codificador de $K$ capas una propuesta para la función de pérdida de regularización sería la siguiente:

$$L_{sparsity}(\theta, \phi) = \mathbb{E}_{x \sim \mu_X} \Big[ \sum_{k \in 1:K} \omega_k ||h_k|| \Big]$$

donde $h_k$ representa el valor del vector de la función de activación de la capa $k$ y $||\cdot||$ normalmente representa la norma L1 o la norma L2, dando lugar respectivamente a el auto-codificador poco denso L1 o L2.

#### 1.4.2 Auto-codificadores de atenuación de ruido (DAE)

Los auto-codificadores de atenuación de ruido [6] introducen la regularización modificando la función de reconstrucción. Durante el entrenamiento, la entrada se distorsiona y se intenta recuperar la información original.

El proceso de distorsión se hace mediante una función $T : \mathcal{X} \to \mathcal{X}$ que a partir de una señal de entrada $x \in \mathcal{X}$ la transforma en una versión distorsionada $T(x)$. El valor de $T$ se selecciona aleatoriamente con una distribución de probabilidad $\mu_T$.

La función de optimización para el caso del DAE es:

$$\min_{\theta, \phi} L(\theta, \phi) = \mathbb{E}_{x \sim \mu_X, T \sim \mu_T}[d(x, (D_\theta \circ E_\phi \circ T)(x))]$$

Funciones típicas utilizadas para la adición de ruido incluyen el ruido isotrópico gaussiano, ruido mediante enmascaramiento aleatorio de parte de la entrada o fijar a valores aleatorios parte de los elementos de la entrada.

#### 1.4.3 Auto-codificador contractivo

Un auto-codificador contractivo [7] añade una pérdida de regularización a la pérdida estándar del auto-codificador:

$$\min_{\theta,\phi} L(\theta,\phi) + \lambda L_{contractive}(\theta,\phi)$$

donde $\lambda$ regula la importancia de la regularización contractiva, que se define como la norma de Frobenius [8] esperada del jacobiano de las activaciones del codificador respecto a la entrada:

$$L_{contractive}(\theta,\phi) = \mathbb{E}_{x \sim \mu_{ref}} ||\nabla_x E_\phi(x)||_F^2$$

Sabemos que $||E_\phi(x+\delta x) - E_\phi(x)||_2 \leq ||\nabla x E_\phi(x)||_F ||\delta x||_2$ para cualquier entrada $x \in \mathcal{X}$ y pequeña variación $\delta x$ sobre ella. Por lo tanto, si $||\nabla_x E_\phi(x)||_F^2$ es pequeño, significa que valores alrededor de la entrada se asignan también a valores situados en un entorno próximo de su representación latente. Esta es una propiedad deseada, ya que significa que una pequeña variación de la entrada conduce a una pequeña variación, tal vez incluso cero, de su valor latente. Esto hace, por ejemplo, que dos imágenes puedan verse iguales incluso si no son exactamente iguales.

El DAE puede entenderse como un límite infinitesimal de CAE: en el límite del ruido de entrada gaussiano, los DAE hacen que la función de reconstrucción resista las perturbaciones de entrada, mientras que el CAE hace que las características extraídas resistan perturbaciones de entrada infinitesimales.

### 1.5 Aplicaciones típicas

Las aplicaciones típicas de los auto-codificadores descritos en esta sección incluyen las siguientes:

- Compresión: utilización de un auto-codificador para representar la señal de entrada con otra de dimensión menor. A efectos comparativos, cuando un auto-codificador usa una función lineal como codificador y decodificador, MSE como función de pérdida y los datos de entrada se normalizan de acuerdo a $\hat{x}_{i,j} = \frac{1}{\sqrt{M}}\left(x_{i,j} - \frac{1}{M}\sum_{k=1}^{M} x_{k,j}\right)$ entonces es equivalente a un PCA [9].
- Clasificación mediante el uso de atributos latentes: los atributos del espacio latente pueden ser utilizados para clasificar imágenes, por ejemplo, utilizando k-NN para llevar a cabo la separación (supervisada o no supervisada) [10].
- Detección de anomalías: se entrena un auto-codificador en un conjunto de datos que contiene un conjunto representativo de las imágenes/datos que se quieren reconocer como válidas para luego calcular el error de reconstrucción de las imágenes/datos anómalas y compararlo con los errores de las imágenes/datos correctas. Si ambos conjuntos de imágenes/datos son suficientemente distintos, el error de reconstrucción de las anómalas será superior al de las correctas. Fijando el umbral en el punto correcto será posible detectar aquellas que sean anómalas pues su error de reconstrucción debería ser más alto [11], [12], [13].
- Atenuación de ruido en imágenes: el auto-codificador se utiliza para corregir errores en la imagen de entrada [14], [15].

## 2 Auto-codificadores variacionales (VAEs)

### 2.1 Introducción

Los VAEs (fig. 2.1) son una tipología de modelos gráficos probabilísticos basados en redes neuronales. Presentan una estructura similar a la de los auto-codificadores introducidos en el capítulo anterior. Permiten realizar inferencia eficiente en la presencia de variables latentes continuas (y/o discretas), con distribuciones intratables del posterior y utilizando conjuntos de datos de gran tamaño.

Los VAEs son capaces de generar nuevos datos similares a los de la distribución de origen. Esto se consigue mediante la codificación, en forma de variables latentes, de las características definitorias de la distribución de probabilidad representada por la muestra poblacional del conjunto de datos original. A partir del modelo aprendido, son capaces de generar nuevas instancias que aproximan a dicha distribución.

Una red codificadora (también llamada red de reconocimiento) transforma la distribución de origen a un espacio latente estocástico representado por una distribución más sencilla (por ejemplo una distribución normal). A partir de la distribución latente se generan los datos de reconstrucción mediante el uso de una red decodificadora (o red de generación).

Al final del entrenamiento, si este ha sido efectivo, será posible generar nuevas instancias mediante el decodificador, transformando las muestras del espacio oculto al espacio de expresión. El codificador actúa como un modelo de reconocimiento, mientras que el decodificador como uno de generación. El modelo de reconocimiento proporciona al generador una aproximación del posterior sobre el espacio latente, que es utilizado para optimizar los parámetros de ambas redes mediante estimación por máxima verosimilitud. Según la regla de Bayes, el modelo de reconocimiento aproxima a la inversa del modelo generativo.

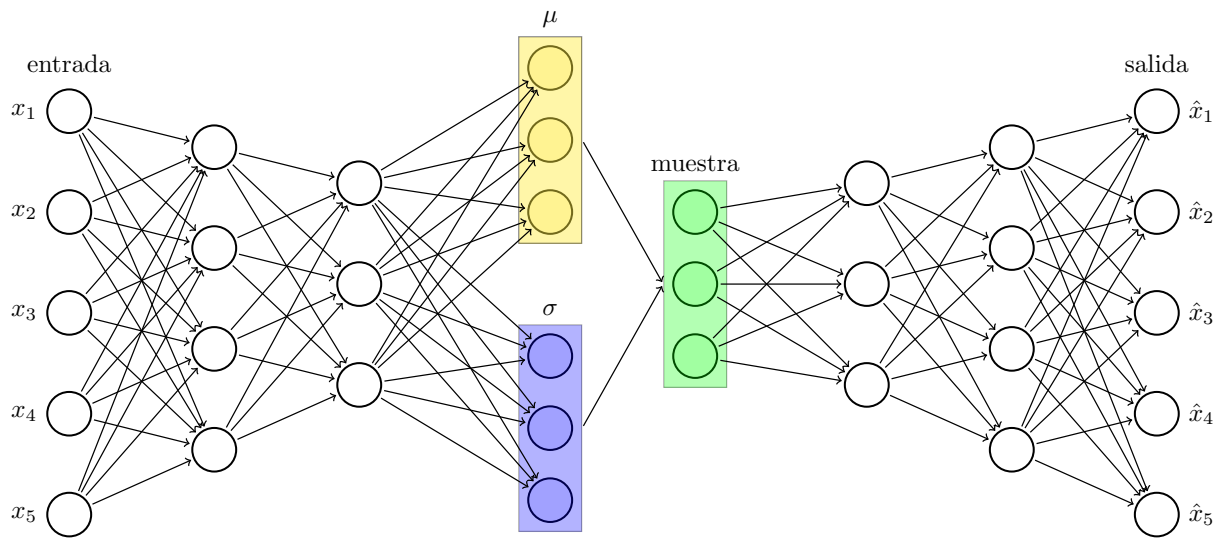

Figura 2: Diagrama representativo de la arquitectura VAE

## 2.2 Modelos probabilísticos

Los modelos probabilísticos son una representación matemática de fenómenos afectos de incertidumbre. Estos quedan definidos mediante distribuciones de probabilidad. Suponiendo que $p^*(\boldsymbol{x})$ represente la distribución de probabilidad original, podemos intentar aproximarla mediante un modelo paramétrico $p_\theta(\boldsymbol{x})$ tal que $p_\theta(\boldsymbol{x}) \approx p^*(\boldsymbol{x})$. Es de interés que $p_\theta(\boldsymbol{x})$ sea capaz de adaptarse a las características de los datos de origen así como de incorporar información que conozcamos a priori, como por ejemplo, el conocimiento procedente de distribuciones de probabilidad condicionadas.

A veces puede suceder que no interese modelizar $p_\theta(\boldsymbol{x})$ sino que nos sea suficiente con modelizar $p_\theta(\boldsymbol{y}|\boldsymbol{x})$ con el objetivo de conseguir $p_\theta(\boldsymbol{y}|\boldsymbol{x}) \approx p^*(\boldsymbol{y}|\boldsymbol{x})$. Para simplificar la notación, en este capítulo nos referiremos a la modelización incondicional, aunque lo explicado aquí también aplique a la condicional.

Los modelos gráficos probabilísticos (PGM) permiten representar a la distribución conjunta de probabilidad atendiendo a las interconexiones causales existentes entre las distintas variables.

$$p_\theta(x_1,...,x_M) = \prod_{j=1}^{M} p_\theta(x_j|Pa(x_j))$$

donde $Pa(x_j)$ representa a los nodos padres de $x_j$. Para los nodos sin padres, la probabilidad referida será incondicionada.

$p_\theta(x_j|Pa(x_j))$ puede ser definida de distintos modos, a saber, como una tabla de valores, como un modelo lineal o también como un red neuronal. En este último caso $p_\theta(x|Pa(\boldsymbol{x})) = p_\theta(\boldsymbol{x}|\eta)$, donde $\eta = RedNeuronal(Pa(\boldsymbol{x}))$. Éstos últimos son modelos paramétricos muy versátiles que permiten modelizar funciones complejas y son una muy buena herramienta para modelizar distribuciones optimizables matemáticamente.

### 2.3 Optimización de modelos probabilísticos

Sea un conjunto de datos $\mathcal{D} = \{\boldsymbol{x}^{(1)}, ..., \boldsymbol{x}^{(N)}\} \equiv \{\boldsymbol{x}^{(i)}\}_{i=1}^{N} = \boldsymbol{x}^{(1:N)}$, donde $\boldsymbol{x}^{(i)}$ son muestras, independientes e idénticamente distribuidas (i.i.d), pertenecientes a una distribución subyacente. El modelo paramétrico que aproxima la distribución conjunta de probabilidad, será de la forma $p_\theta(\mathcal{D}) = \prod_{\boldsymbol{x} \in \mathcal{D}} p_\theta(\boldsymbol{x})$ que en su versión logarítmica se expresa como $\log p_\theta(\mathcal{D}) = \sum_{\boldsymbol{x} \in \mathcal{D}} \log p_\theta(\boldsymbol{x})$. La forma típica de calcular $\theta$ es optimizar esta función, esto es, la estimación por máxima verosimilitud (MLE) de las instancias pertenecientes al conjunto de datos $\mathcal{D}$. El método MLE equivale a la minimización de la divergencia de Kullback-Leibler entre la distribución de los datos y la del modelo [16]. Disponiendo de la función de optimización MLE ya es posible calcular los parámetros $\theta$ de la red neural utilizando los métodos de entrenamiento convencionales, por ejemplo, el descenso de gradiente estocástico.

### 2.4 Inferencia variacional (VI)[17]

#### 2.4.1 Variables observables vs ocultas

Consideramos a una *variable observable*, cuando forma parte de aquellas de las que disponemos de su valor en los puntos que forman parte del conjunto de datos. Una *variable oculta o latente* es aquella de la que no tenemos información directa a través del conjunto de entrenamiento conocido. Esta formalización permite conectar la información percibida a través de variables observadas con las abstracciones propias de los modelos interpretativos, que serían las variables ocultas o latentes que queremos calcular. Así, por ejemplo, en un conjunto de datos formado por imágenes, las variables observables serían los píxeles de los que disponemos de información, mientras que las ocultas de interés podrían ser las abstracciones en forma de modelos del mundo que hacemos los seres inteligentes, por ejemplo, los distintos tipos de objetos presentes en la imagen.

#### 2.4.2 Formalización matemática

Supongamos que partimos de una distribución de probabilidad representada por un conjunto de datos $\mathcal{D}$, con $\boldsymbol{x} \in \mathcal{D}$, de la que queremos inferir una serie de propiedades ocultas (p.e. abstracciones), representadas por un conjunto de variables, originalmente desconocidas, que llamaremos $\boldsymbol{z}$. Sea $p(\boldsymbol{x}, \boldsymbol{z})$ la distribución conjunta de las variables observables y ocultas.

En estos casos, nos interesa realizar la inferencia acerca de las variables desconocidas, esto es, calcular $p(\boldsymbol{z}|\boldsymbol{x})$. Sabemos que este valor se puede representar como:

$$p(\boldsymbol{z}|\boldsymbol{x}) = \frac{p(\boldsymbol{x}, \boldsymbol{z})}{p(\boldsymbol{x})} = \frac{p(\boldsymbol{x}|\boldsymbol{z})p(\boldsymbol{z})}{p(\boldsymbol{x})}$$

Calcular $p(\boldsymbol{z}|\boldsymbol{x})$ a partir de la igualdad bayesiana puede tener su dificultad pues habitualmente alguna de las variables suele resultar intratable computacionalmente.

La inferencia variacional trata de resolver el problema mediante una aproximación, esto es, restringiendo el espacio de búsqueda a una familia de funciones sencillas, predefinida de origen, que llamaremos familia de funciones variacional $q_\phi(\boldsymbol{z}|\boldsymbol{x})$. Es común, aunque no requisito, que la familia escogida sea la de funciones gaussianas.

A partir de aquí, se convierte el problema original en uno de optimización consistente en encontrar los parámetros $\phi$ óptimos restringidos a la familia de funciones $q_\phi(\boldsymbol{z}|\boldsymbol{x})$ que minimizan la distancia existente entre las distribuciones $p(\boldsymbol{z}|\boldsymbol{x})$ y $q_\phi(\boldsymbol{z}|\boldsymbol{x})$. Habitualmente, la función de medida de distancia entre distribuciones escogida es la divergencia de Kullback-Leibler:

$$D_{KL}(q_\phi(\boldsymbol{z}|\boldsymbol{x})||p(\boldsymbol{z}|\boldsymbol{x})) = \mathbb{E}_{z \sim q}\big[\log \frac{q_\phi(\boldsymbol{z}|\boldsymbol{x})}{p(\boldsymbol{z}|\boldsymbol{x})}\big] = \sum_{x \in \mathcal{X}} q_\phi(\boldsymbol{z}|\boldsymbol{x}) \log \Big(\frac{q_\phi(\boldsymbol{z}|\boldsymbol{x})}{p(\boldsymbol{z}|\boldsymbol{x})}\Big)$$

En este caso, el objetivo de la inferencia variacional es encontrar:

$$\hat{\theta} = arg \min_{\phi} D_{KL}(q_\phi(\boldsymbol{z}|\boldsymbol{x})||p(\boldsymbol{z}|\boldsymbol{x}))$$

Existen muchos algoritmos distintos para abordar este problema, cada uno con sus ventajas e inconvenientes que dependen de la naturaleza de $p(\boldsymbol{x}, \boldsymbol{z})$ y de la familia subrogada de $q$.

La inferencia variacional define una estrategia genérica para abordar el problema, pero no un algoritmo concreto para conseguirlo.

### 2.4.3 Límite inferior de evidencia (ELBO)

Llegados a este punto, vamos a definir una variable nueva partiendo de $p(\boldsymbol{x})$ que nos será de utilidad para facilitar posteriormente la estrategia de optimización.

Para ello nos serviremos de una la propiedad de las funciones convexas, denominada *desigualdad de Jensen* que dice lo que sigue: sea $f$ una función convexa en el dominio $\mathcal{X}$, entonces $\mathbb{E}[f(x)] \leq f(\mathbb{E}[x]), \forall x \in \mathcal{X}$. A efectos de nuestro caso concreto, si tomamos a $f$ como la función logarítmica, tenemos que $\mathbb{E}[\log(x)] \leq \log(\mathbb{E}[x])$ para todo el dominio de la función.

Utilizando dicha propiedad, simplificamos la expresión de $\log p(\boldsymbol{x})$, creando un límite inferior:

$$\begin{aligned}
\log p(\boldsymbol{x}) &= \log \int p(\boldsymbol{x}, \boldsymbol{z})\, d\boldsymbol{z} = \\
&= \log \int p(\boldsymbol{x}, \boldsymbol{z}) \frac{q_\phi(\boldsymbol{z}|\boldsymbol{x})}{q_\phi(\boldsymbol{z}|\boldsymbol{x})}\, d\boldsymbol{z} = \\
&= \log \mathbb{E}_{z \sim q} \frac{p(\boldsymbol{x}, \boldsymbol{z})}{q_\phi(\boldsymbol{z}|\boldsymbol{x})}\, d\boldsymbol{z} \geq \\
&\geq \mathbb{E}_{z \sim q} \log \frac{p(\boldsymbol{x}, \boldsymbol{z})}{q_\phi(\boldsymbol{z}|\boldsymbol{x})}\, d\boldsymbol{z} = \\
&= \mathbb{E}_{z \sim q} \log p(\boldsymbol{x}, \boldsymbol{z}) - \mathbb{E}_{z \sim q} \log q_\phi(\boldsymbol{z}|\boldsymbol{x}) \\
&= ELBO = \mathcal{L}_{\theta, \phi}(\boldsymbol{x})
\end{aligned}$$

A dicho límite inferior lo llamamos *límite inferior de evidencia*, $ELBO$ en inglés o $\mathcal{L}_{\theta,\phi}(\boldsymbol{x})$ en su expresión matemática en este texto.

El ELBO también se puede expresar de forma alternativa como:

$$\begin{aligned}
\mathcal{L}_{\theta,\phi}(\boldsymbol{x}) &= \mathbb{E}_{z \sim q} \log p(\boldsymbol{x}, \boldsymbol{z}) - \underbrace{\mathbb{E}_{z \sim q} \log q_\phi(\boldsymbol{z}|\boldsymbol{x})}_{\text{Entropía de q}} \\
&= \underbrace{\mathbb{E}_{z \sim q} \log p(\boldsymbol{x}|\boldsymbol{z})}_{\text{- Pérdida de reconstrucción}} - \underbrace{D_{KL}(q_\phi(\boldsymbol{z}|\boldsymbol{x})||p(\boldsymbol{z}|\boldsymbol{x}))}_{\text{KL entre q y el prior}}
\end{aligned}$$

### 2.4.4 ELBO como medio de optimizar $q_\phi$

A continuación demostraremos que $D_{KL}(q_\phi(\boldsymbol{z}|\boldsymbol{x})||p(\boldsymbol{z}|\boldsymbol{x})) = \log p(\boldsymbol{x}) - \mathcal{L}_{\theta,\phi}(\boldsymbol{x})$, esto es, que maximizando el ELBO estamos minimizando también la distancia entre las distribuciones $p(\boldsymbol{z}|\boldsymbol{x})$ y $q_\phi(\boldsymbol{z}|\boldsymbol{x})$ y, por lo tanto, consiguiendo el objetivo buscado de aproximar $p(\boldsymbol{z}|\boldsymbol{x})$ con la función más sencilla $q_\phi(\boldsymbol{z}|\boldsymbol{x})$.

Demostración:

$$\begin{aligned}
D_{KL}(q_\phi(\boldsymbol{z}|\boldsymbol{x})||p(\boldsymbol{z}|\boldsymbol{x})) &= \mathbb{E}_{z \sim q}\left[\log \frac{q_\phi(\boldsymbol{z}|\boldsymbol{x})}{p(\boldsymbol{z}|\boldsymbol{x})}\right] = \\
&= \mathbb{E}_{z \sim q}\left[\log q_\phi(\boldsymbol{z}|\boldsymbol{x})\right] - \mathbb{E}_{z \sim q}\left[\log p(\boldsymbol{z}|\boldsymbol{x})\right] = \\
&= \mathbb{E}_{z \sim q}\left[\log q_\phi(\boldsymbol{z}|\boldsymbol{x})\right] - \mathbb{E}_{z \sim q}\left[\log \frac{p(\boldsymbol{x}, \boldsymbol{z})}{p(\boldsymbol{x})}\right] = \\
&= \mathbb{E}_{z \sim q}\left[\log q_\phi(\boldsymbol{z}|\boldsymbol{x})\right] - \mathbb{E}_{z \sim q}\left[\log p(\boldsymbol{x}, \boldsymbol{z})\right] + \mathbb{E}_{z \sim q}\left[\log p(\boldsymbol{x})\right] = \\
&= \log p(\boldsymbol{x}) - \left(\mathbb{E}_{z \sim q}[\log p(\boldsymbol{x}, \boldsymbol{z})] - \mathbb{E}_{z \sim q}[\log q_\phi(\boldsymbol{z}|\boldsymbol{x})]\right) = \\
&= \log p(\boldsymbol{x}) - \mathcal{L}_{\theta,\phi}(\boldsymbol{x})
\end{aligned}$$

Sabemos que el ELBO es un límite inferior de la evidencia, esto es,

$$\mathcal{L}_{\theta,\phi}(\boldsymbol{x}) = \log p_\theta(\boldsymbol{x}) - D_{KL}(q_\phi(\boldsymbol{z}|\boldsymbol{x})||p(\boldsymbol{z}|\boldsymbol{x})) \leq \log p_\theta(\boldsymbol{x})$$

A partir de esta ecuación podemos entender que maximizando el ELBO con respecto a los parámetros $\phi$ y $\theta$ se consiguen dos objetivos: por un lado maximizar la probabilidad marginal $p_\theta(\boldsymbol{x})$ (nuestro modelo generativo se vuelve mejor) y por otro minimizar la divergencia KL de la aproximación $q_\phi(\boldsymbol{z}|\boldsymbol{x})$ sobre el posterior exacto $p_\theta(\boldsymbol{z}|\boldsymbol{x})$ (la aproximación $q_\phi(\boldsymbol{z}|\boldsymbol{x})$ mejora).

### 2.5 El codificador como forma de aproximar al posterior

Nuestro objetivo es aproximar al posterior, esto es

$$q_\phi(\boldsymbol{z}|\boldsymbol{x}) \approx p_\theta(\boldsymbol{z}|\boldsymbol{x})$$

El modelo de inferencia puede ser un modelo del tipo:

$$q_\phi(\boldsymbol{z}|\boldsymbol{x}) = q_\phi(\boldsymbol{z}_1, ..., \boldsymbol{z}_M|\boldsymbol{x}) = \prod_{j=1}^{M} q_\phi(\boldsymbol{z}_j|Pa(\boldsymbol{z}_j), \boldsymbol{x})$$

donde $Pa(\boldsymbol{z}_j)$ es el conjunto de nodos precedentes a la variable $\boldsymbol{z}_j$. La distribución $q_\phi(\boldsymbol{z}|\boldsymbol{x})$ puede ser igualmente modelada como una red neuronal profunda. De cara a encontrar los parámetros óptimos de la red neuronal, necesitamos definir una función objetivo tratable.

Utilizando la inferencia variacional definida en el apartado anterior, restringiendo el espacio de búsqueda a la familia de funciones gaussianas, la función a optimizar tendría la forma siguiente:

$$\begin{aligned}(\boldsymbol{\mu}, \log \boldsymbol{\sigma}) &= RedNeuralCodificador_\phi(\boldsymbol{x}) \\ q_\phi(\boldsymbol{z}|\boldsymbol{x}) &= \mathcal{N}(\boldsymbol{z}; \boldsymbol{\mu}, diag(\boldsymbol{\sigma}))\end{aligned}$$

Esto es, la red neuronal serviría para calcular, para cada uno de los componentes de $\boldsymbol{z}$, la media $\mu$ y desviación estándar $\sigma$ de una distribución normal que determinarían $q_\phi$, siendo $\phi$ realmente los parámetros que definen a $\boldsymbol{\mu}$ y $\boldsymbol{\sigma}$.

En los métodos tradicionales de inferencia variacional los parámetros se optimizan de manera separada para cada punto de datos. En el caso del auto-codificador variacional, los parámetros se suponen comunes a todo el conjunto de datos. A esta aproximación al problema se la denomina *inferencia variacional amortizada* [18] y permite la eliminación del bucle de optimización por punto de datos. Con esta suposición, implícitamente se está suponiendo que la expresividad de la red neuronal es suficiente para codificar la dependencia respecto a $\boldsymbol{x}$. Esto permite aprovechar la eficiencia del descenso de gradiente estocástico.

### 2.6 Optimización del ELBO mediante gradiente estocástico

Para poder utilizar al ELBO como función objetivo, es necesario derivar los gradientes respecto a $\theta$ y $\phi$. Dado un conjunto i.i.d. de datos $\mathcal{D}$ el ELBO se puede representar como:

$$\mathcal{L}_{\theta,\phi}(\mathcal{D}) = \sum_{x \in \mathcal{D}} \mathcal{L}_{\theta,\phi}(\boldsymbol{x})$$

Normalmente, este objetivo suele ser intratable, aunque suele ser posible, tal como veremos a continuación, obtener buenos estimadores no sesgados que permitan llevar a cabo SGD.

El gradiente respecto a $\theta$ suele ser fácil de obtener:

$$\begin{aligned}\nabla_\theta \mathcal{L}_{\theta,\phi}(\boldsymbol{x}) &= \nabla_\theta \mathbb{E}_q[\log p_\theta(\boldsymbol{x}, \boldsymbol{z})] - \log q_\phi(\boldsymbol{z}|\boldsymbol{x})] \\ &= \mathbb{E}_q[\nabla_\theta(\log p_\theta(\boldsymbol{x}, \boldsymbol{z})) - \log q_\phi(\boldsymbol{z}|\boldsymbol{x})] \\ &= \mathbb{E}_q[\nabla_\theta(\log p_\theta(\boldsymbol{x}, \boldsymbol{z}))] \\ &\simeq \nabla_\theta \log p_\theta(\boldsymbol{x}, \boldsymbol{z})\end{aligned}$$

La última linea corresponde con el estimador de la esperanza simple de Monte Carlo [19].

No es tan sencillo para el caso de $\nabla_\phi$, pues tal como veremos en la derivación presentada a continuación, $\mathbb{E}_q$ depende de $\phi$.

$$\begin{aligned}\nabla_\phi \mathcal{L}_{\theta,\phi}(\boldsymbol{x}) &= \nabla_\theta \mathbb{E}_q[\log p_\theta(\boldsymbol{x}, \boldsymbol{z})] - \log q_\phi(\boldsymbol{z}|\boldsymbol{x})] \\ &\neq \mathbb{E}_q \nabla_\theta[\log p_\theta(\boldsymbol{x}, \boldsymbol{z})] - \log q_\phi(\boldsymbol{z}|\boldsymbol{x})]\end{aligned}$$

Para poder conseguir simplificar la esperanza, es necesario deshacer la dependencia de $\phi$. Esto es posible hacerlo realizando un cambio de variable que desligue el muestreo de $\phi$, esto es, convertir $\boldsymbol{z} \sim q_\phi(\boldsymbol{z}|\boldsymbol{x})$ en $\boldsymbol{\epsilon} \sim p(\boldsymbol{\epsilon})$ y expresar $\boldsymbol{z}$ como $\boldsymbol{z} = \boldsymbol{g}(\boldsymbol{\epsilon}, \phi, \boldsymbol{x})$.

Realizando dicho cambio de variable, la esperanza puede ser expresada como sigue:

$$\mathbb{E}_{\boldsymbol{z} \sim q_\phi(\boldsymbol{z}|\boldsymbol{x})}[f(\boldsymbol{z})] = \mathbb{E}_{\boldsymbol{\epsilon} \sim p(\boldsymbol{\epsilon})}[f(\boldsymbol{z})]$$
$$f(\boldsymbol{z}) = \log p_\theta(\boldsymbol{x}, \boldsymbol{z}) - \log q_\phi(\boldsymbol{z}|\boldsymbol{x})$$
$$\boldsymbol{z} = \boldsymbol{g}(\boldsymbol{\phi}, \boldsymbol{x}, \boldsymbol{\epsilon})$$

A partir de este cambio de variable, ya es posible derivar el estimador de Monte Carlo:

$$\begin{aligned} \nabla_\phi \mathbb{E}_{\boldsymbol{z} \sim q_\phi(\boldsymbol{z}|\boldsymbol{x})}[f(\boldsymbol{z})] &= \nabla_\phi \mathbb{E}_{\boldsymbol{\epsilon} \sim p(\boldsymbol{\epsilon})}[f(\boldsymbol{z})] \\ &= \mathbb{E}_{\boldsymbol{\epsilon} \sim p(\boldsymbol{\epsilon})} \nabla_\phi [f(\boldsymbol{z})] \\ &\simeq \nabla_\phi [f(\boldsymbol{z})] \end{aligned}$$

Esta reparametrización $\boldsymbol{z} = \boldsymbol{g}(\boldsymbol{\phi}, \boldsymbol{x}, \boldsymbol{\epsilon})$ también permite expresar al ELBO como sigue:

$$\begin{aligned} \mathcal{L}_{\theta,\phi}(\boldsymbol{x}) &= \mathbb{E}_{\boldsymbol{z} \sim q_\phi(\boldsymbol{z}|\boldsymbol{x})}[\log p_\theta(\boldsymbol{x}, \boldsymbol{z}) - \log q_\phi(\boldsymbol{z}|\boldsymbol{x})] \\ &= \mathbb{E}_{\boldsymbol{\epsilon} \sim p_\epsilon}[\log p_\theta(\boldsymbol{x}, \boldsymbol{z}) - \log q_\phi(\boldsymbol{z}|\boldsymbol{x})] \end{aligned}$$

A partir de esta expresión podemos derivar un estimador simple de Monte Carlo $\tilde{\mathcal{L}}_{\theta,\phi}(\boldsymbol{x})$ del ELBO de un punto utilizando una única muestra de ruido $\boldsymbol{\epsilon}$ de $p(\boldsymbol{\epsilon})$:

$$\boldsymbol{\epsilon} \sim p(\boldsymbol{\epsilon})$$
$$\boldsymbol{z} = \boldsymbol{g}(\boldsymbol{\phi}, \boldsymbol{x}, \boldsymbol{\epsilon})$$
$$\tilde{\mathcal{L}}_{\theta,\phi}(\boldsymbol{x}) = \log p_\theta(\boldsymbol{x}, \boldsymbol{z}) - \log q_\phi(\boldsymbol{z}|\boldsymbol{x})$$

Este estimador puede ser fácilmente calculado mediante librerías de cómputo con diferenciación automática de forma que sea sencillo calcular las derivadas respecto a los parámetros $\theta$ y $\phi$. El valor de $\nabla_\phi \tilde{\mathcal{L}}_{\theta,\phi}(\boldsymbol{x})$ se puede utilizar para optimizar el ELBO mediante descenso de gradiente [20]. En el algoritmo 1 describe la operativa de optimización.

**Algorithm 1** Optimización estocástica del ELBO. A este procedimiento de optimización se le denomina AEVB, i.e. *auto-codificación bayesiana variacional*

**Entrada:** $\mathcal{D}$, conjunto de datos de entrenamiento
**Parámetros:** $q_\phi(\boldsymbol{z}|\boldsymbol{x})$, modelo de inferencia
**Parámetros:** $p_\theta(\boldsymbol{x}, \boldsymbol{z})$, modelo generativo
1) $(\boldsymbol{\theta}, \boldsymbol{\phi}) \leftarrow$ Inicializacion
2) **mientras** SGD no converja **hacer**
3)    $\mathcal{M} \sim \mathcal{D}$ (seleccionar minibatch)
4)    $\boldsymbol{\epsilon} \sim p(\boldsymbol{\epsilon})$ (ruido aleatorio disitinto para cada muestra de $\mathcal{M}$)
5    Calcular $\tilde{\mathcal{L}}_{\theta,\phi}(\mathcal{M}, \boldsymbol{\epsilon})$ y sus gradientes $\nabla_{\theta,\phi} \tilde{\mathcal{L}}_{\theta,\phi}(\mathcal{M}, \boldsymbol{\epsilon})$
6)    Actualizar $\theta$ y $\phi$ utilizando el optimizador SGD
7) **fin**

El gradiente utilizado es un estimador no sesgado del gradiente del ELBO. La media del muestreo sobre $\boldsymbol{\epsilon} \sim p(\boldsymbol{\epsilon})$ equivale al gradiente del ELBO sobre el punto considerado.

### 2.7 Cálculo de $q_\phi$

Para obtener el estimador del ELBO es necesario calcular la densidad $q_\phi(\boldsymbol{z}|\boldsymbol{x})$. El cálculo de este valor es sencillo si se escoge una transformación adecuada $\boldsymbol{g}$. La densidad $p(\boldsymbol{\epsilon})$ es conocida, pues corresponde con la densidad de la distribución escogida de antemano.

Si $\boldsymbol{g}(\cdot)$ es una función invertible, entonces las funciones de densidad de probabilidad de $\boldsymbol{\epsilon}$ y $\boldsymbol{z}$ están relacionadas por la siguiente expresión [2]:

$$\log q_\phi(\boldsymbol{z}|\boldsymbol{x}) = \log p(\boldsymbol{\epsilon}) - \log d_\phi(\boldsymbol{x}, \boldsymbol{\epsilon})$$

---
[2] Para prueba consultar apartado 1.6.10 en [21]

donde:

$$\log d_\phi(\boldsymbol{x}, \boldsymbol{\epsilon}) = \log \left| det\left(\frac{\partial \boldsymbol{z}}{\partial \boldsymbol{\epsilon}}\right) \right|$$

siendo

$$\frac{\partial \boldsymbol{z}}{\partial \boldsymbol{\epsilon}} = \frac{\partial(z_1, ..., z_k)}{\partial(\epsilon_1, ..., \epsilon_k)} = \begin{pmatrix} \frac{\partial z_1}{\partial \epsilon_1} & \cdots & \frac{\partial z_1}{\partial \epsilon_k} \\ \vdots & \ddots & \vdots \\ \frac{\partial z_k}{\partial \epsilon_1} & \cdots & \frac{\partial z_k}{\partial \epsilon_k} \end{pmatrix}$$

### 2.8 Posteriores Gaussianos

Es común escoger un codificador Gaussiano $q_\phi(\boldsymbol{z}|\boldsymbol{x}) = \mathcal{N}(\boldsymbol{z}; \boldsymbol{\mu}, diag(\boldsymbol{\sigma}^2))$:

$$(\boldsymbol{\mu}, \log \boldsymbol{\sigma}) = RedNeuralCodificador_\phi(\boldsymbol{x})$$
$$q_\phi(\boldsymbol{z}|\boldsymbol{x}) = \prod_i q_\phi(z_i|\boldsymbol{x}) = \prod_i \mathcal{N}(z_i; \mu_i, \sigma_i^2)$$

Que después de reparametrizar queda como:

$$\boldsymbol{\epsilon} \sim \mathcal{N}(0, \boldsymbol{I})$$
$$(\boldsymbol{\mu}, \log \boldsymbol{\sigma}) = RedNeuralCodificador(\boldsymbol{x})$$
$$\boldsymbol{z} = \boldsymbol{\mu} + \boldsymbol{\sigma} \odot \boldsymbol{\epsilon}$$

donde $\odot$ representa el producto elemento a elemento. La transformación Jacobiana de $\boldsymbol{\epsilon}$ a $\boldsymbol{z}$ es:

$$\frac{\partial \boldsymbol{z}}{\partial \boldsymbol{\epsilon}} = diag(\boldsymbol{\sigma})$$

El logaritmo del determinante del Jacobiano, siguiendo las ecuaciones del apartado anterior para el caso gaussiano se puede expresar como:

$$\log d_\phi(\boldsymbol{x}, \boldsymbol{\epsilon}) = \log \left| det\left(\frac{\partial \boldsymbol{z}}{\partial \boldsymbol{\epsilon}}\right) \right| = \sum_i \log \sigma_i$$

y la densidad del posterior es:

$$\log q_\phi(\boldsymbol{z}|\boldsymbol{x}) = \sum_i \log \mathcal{N}(\epsilon_i; 0, 1) - \log \sigma_i$$

cuando $\boldsymbol{z} = g(\boldsymbol{\epsilon}, \phi, \boldsymbol{x})$.

### 2.9 Posterior Gaussiano con matriz de covarianza general

El posterior gaussiano factorizado puede ser extendido al caso gaussiano con covarianza completa:

$$q_\phi(\boldsymbol{z}|\boldsymbol{x}) = \mathcal{N}(\boldsymbol{z}; \boldsymbol{\mu}, \boldsymbol{\Sigma})$$

La reparametrización vendría dada por

$$\boldsymbol{\epsilon} \sim \mathcal{N}(0, \boldsymbol{I})$$
$$\boldsymbol{z} = \boldsymbol{\mu} + \boldsymbol{L}\boldsymbol{\epsilon}$$

donde $\boldsymbol{L}$ es la matriz triangular superior (o inferior) con valores no nulos en la diagonal resultante de realizar la descomposición de Cholesky, esto es $\boldsymbol{\Sigma} = \boldsymbol{L}\boldsymbol{L}^T$. Se propone este cambio de variable ya que de esta forma el cálculo del Jacobiano es trivial: $\frac{\partial \boldsymbol{z}}{\partial \boldsymbol{\epsilon}} = \boldsymbol{L}$, resulta ser el producto de los elementos de la diagonal:

$$\log d_\phi(\boldsymbol{x}, \boldsymbol{\epsilon}) = \log \left| det\left(\frac{\partial \boldsymbol{z}}{\partial \boldsymbol{\epsilon}}\right)\right| = \sum_i \log L_{ii}$$

Y la densidad logarítmica del posterior se puede representar como:

$$\log q_\phi(\boldsymbol{z}|\boldsymbol{x}) = \log p(\epsilon) - \sum_i \log |L_{ii}|$$

Calculemos ahora la varianza de $\boldsymbol{z}$ para demostrar que se corresponde con la resultante de la descomposición de Cholesky:

$$\begin{aligned}\boldsymbol{\Sigma}(\boldsymbol{z}) &= \mathbb{E}\big[(\boldsymbol{z} - \mathbb{E}[\boldsymbol{z}])(\boldsymbol{z} - \mathbb{E}[\boldsymbol{z}])^T\big] \\ &= \mathbb{E}\big[\boldsymbol{L}\boldsymbol{\epsilon}(\boldsymbol{L}\boldsymbol{\epsilon})^T\big] = \boldsymbol{L}\mathbb{E}\big[\boldsymbol{\epsilon}\boldsymbol{\epsilon}^T\big]\boldsymbol{L}^T = \boldsymbol{L}\boldsymbol{L}^T\end{aligned}$$

Hacer notar que $\mathbb{E}\big[\boldsymbol{\epsilon}\boldsymbol{\epsilon}^T\big] = \boldsymbol{I}$ ya que $\boldsymbol{\epsilon} \sim \mathcal{N}(0, \boldsymbol{I})$.

Una manera de calcular los parámetros de la función $q_\phi$ sería utilizar una red neuronal para predecir $\boldsymbol{\mu}$, $\boldsymbol{\Sigma}$. El problema de esta aproximación es que para calcular $\boldsymbol{L}$ a través de la descomposición de Cholesky, incurriríamos en un coste computacional $\mathcal{O}(n^3)$ para calcular $\boldsymbol{L}$. Una forma alternativa más económica computacionalmente es calcular directamente $\boldsymbol{L}$, por ejemplo, de la siguiente forma:

$$\begin{aligned}(\boldsymbol{\mu}, \log \boldsymbol{\sigma}, \boldsymbol{L}') &\leftarrow RedNeuronalCodificador_\phi(\boldsymbol{x}) \\ \boldsymbol{L} &\leftarrow \boldsymbol{L}_{mask} \cdot \boldsymbol{L}' + diag(\boldsymbol{\sigma})\end{aligned}$$

La red neuronal predice $\mu$, $\log \sigma$ y una matriz $\boldsymbol{L}'$. De esta última únicamente se consideran los elementos de debajo de la diagonal. Esto se hace aplicando una máscara $\boldsymbol{L}_{mask}$ para descartar los no necesarios. Este principio de derivación se utiliza en la variación de VAE denominada *flujo auto-regresivo inverso*[22].

### 2.10 Desafíos de la arquitectura

Uno de los problemas típicos al utilizar la optimización estocástica de los auto-codificadores variacionales es que durante el proceso de entrenamiento la red quede atrapada en una zona de equilibrio no óptimo que dificulte el progreso, dando como resultado redes deficientes. Distintas soluciones se han planteado para este problema, que quedan fuera del alcance de un capítulo introductorio a los VAEs. Para el que tenga interés en profundizar sobre este aspecto, puede consultar, por ejemplo [23], [24], [22].

Las características definitorias del método de los VAEs hacen que busquemos una aproximación $q_\phi(\boldsymbol{z}|\boldsymbol{x})$ a $p_\theta(\boldsymbol{z}|\boldsymbol{x})$. Cuando no es posible conseguir una buena aproximación, ya sea porque la función $q_\phi(\boldsymbol{z}|\boldsymbol{x})$ no sea suficientemente expresiva para aproximar a $p_\theta(\boldsymbol{z}|\boldsymbol{x})$, o bien por problemas en la optimización, nos encontremos con la generación de imágenes borrosas típicas de este tipo de arquitecturas.

### 2.11 Posteriores no Gaussianos

Una forma de mejorar el rendimiento de los auto-codificadores variacionales es a través de la mejora de la flexibilidad del modelo de inferencia $q_\phi(\boldsymbol{z}|\boldsymbol{x})$. Mejorando la capacidad expresiva de la función tiene una incidencia directa en el ajuste del límite inferior de evidencia a la probabilidad marginal. Cualquier modificación del modelo de inferencia debe cumplir dos premisas (1) que sea computacionalmente eficiente de calcular y diferenciar, y (2) computacionalmente eficiente de muestrear, pues ambas operaciones deben ser realizadas para cada punto del minibatch, para cada iteración de optimización. También es deseable que el algoritmo sea paralelizable para los elementos de $\boldsymbol{z}$, pues esto permite trabajar con dimensiones altas del vector de variables latentes.

### 2.12 Propuestas de extensión del auto-codificador variacional original

En esta sección enumeramos algunas de las arquitecturas derivadas del auto-codificador variacional original [20] que extienden sus características.

Entre ellas podemos encontrar las que siguen:

- Structured VAE [25], Auxiliary Deep VAE [26], Conditional VAE [27]: Estos tipos de VAE permiten condicionar el aprendizaje de los espacios latentes y la generación a distintos tipos de variables auxiliares.
- Wasserstein autoencoders (WAE-MMD) [28], Sliced-Wasserstein VAE (SWAE) [29]: VAEs que miden la distancia entre distribuciones mediante la distancia Wasserstein en vez de la divergencia Kullback-Leibler.
- $\beta$-VAE [30], [31], [32]: Introduce una constante $\beta$ que se utiliza durante el proceso de entrenamiento para modificar el peso de la divergencia de Kullback-Leibler en la función objetivo. Esto incide en la capacidad de separación de las características representadas por las variables latentes.
- Importance weighted autoencoders (IWAE) [33], MIWAE [34]: Utilizan una estimación más precisa del ELBO basada en el cálculo de varios valores de $\epsilon$ por muestra. Esto permite obtener un ELBO más ajustado a $p(z|x)$.
- Deep feature consistent VAE (DFCVAE) [35]: Añade a la función de pérdida de una VAE convencional, la suma de los MSEs obtenidos de comprarar las activaciones de la imagen original y la imagen reconstruida al paso de una CNN preentrenada. El objetivo es reducir la apariencia borrosa de las imágenes predichas por el VAE.
- MS-SSIM-VAE [36] Utiliza funciones de pérdida basadas en el uso de métricas de similaridad perceptiva (combinación de luminancia, contraste y estructura).
- Categorical VAE [37]: Propone una implementación con el uso de variables latentes categóricas mediante una aproximación diferenciable a su distribución de probabilidad.
- Joint-VAE [38]: Propone el uso combinado de variables latentes continuas y discretas.
- Info-VAE [39]: Propone una función de optimización alternativa para abordar el problema típico de los VAEs cuando el generador obvia parte de la información aprendida para el codificador, generando imágenes subóptimas. Para esto introduce un término el objetivo del cual el maximizar la información mutua entre el espacio latente y el espacio de generación.
- LogCosh-VAE [40]: Propone el uso de una función de pérdida alternativa basada en el logaritmo del coseno hiperbólico para mejorar la calidad visual de las imágenes reconstruidas.
- DIP-VAE [41] Propuesta alternativa para la separación de las variables latentes del VAE para datos no etiquetados.
- Vector Quantized Variational Autoencoder (VQ-VAE) [42]: Utiliza también un espacio latente discreto. Utiliza un "libro de códigos"para almacenar los valores permitidos para el vector del espacio latente. Los vectores presentes el el libro de códigos son aprendidos mediante descenso de gradiente.
- Normalizing Flows [43], [44] se construye una distribución flexible del posterior a través de un procedimiento iterativo. Se comienza con una variable aleatoria con una distribución relativamente simple con una conocida (y computacionalmente económica) función de densidad de probabilidad, para luego derivar por iteración una cadena de transformaciones invertibles parametrizadas de forma que la última iteración tenga una distribución más flexible. Esta operativa no escala bien para espacios dimensionales grandes.
- Inverse autoregressive Flow (IAF) [22] Al igual que los flujos de normalización se empieza con una distribución de probabilidad tratable (p.e. con una gaussiana con covarianza diagonal) seguido de una serie de transformaciones invertibles no lineales de $z$. Se espera que la iteración final tenga una distribución flexible.

En [45] podéis encontrar implementaciones de algunos de estos auto-codificadores variacionales.

## 3   Conclusiones

Los modelos probabilísticos son un pilar fundamental en el campo de la inteligencia artificial moderna. Los modelos probabilísticos con variables latentes permiten asociar variables observables con otras ocultas. Estos pueden utilizarse para descubrir abstracciones, en forma de variables latentes, que sirvan para interpretar las regularidades estadísticas observadas en los datos. Las distribuciones de probabilidad derivadas de relacionar dichas variables con las observables, pueden ser usadas a modo de función objetivo para el diseño de modelos predictivos de los datos procedentes de dichas distribuciones. El problema habitual en estos casos es que las relaciones entre distribuciones suelen implicar cálculos intratables desde un punto de vista computacional. La inferencia variacional (VI) permite convertir estos problemas intratables en otros tratables mediante optimización. La distribución de probabilidad que relaciona variables ocultas con observables, esto es, el posterior $p(oculta|datos)$, se aproxima con una distribución de probabilidad más sencilla (p.e. gaussiana) y se optimiza para que la distancia entre ésta y el posterior real sea lo más pequeña posible. Esto se hace introduciendo unos nuevos modelos paramétricos denominados variacionales. Las redes neuronales, gracias a su alta flexibilidad y expresividad, son un candidato perfecto para su optimización en estas circunstancias.

La inferencia variacional en los VAEs realiza la optimización típicamente a través de un límite inferior de la evidencia (ELBO) mediante optimización por descenso de gradiente estocástico. El VAE resulta ser una combinación de un modelo profundo de variables latentes continuas y un modelo de inferencia asociado. El modelo de inferencia, denominado codificador o modelo de reconocimiento aproxima la distribución del posterior. Tanto el modelo generativo como el de inferencia son modelos probabilísticos basados en redes neuronales. Los parámetros de ambos modelos son optimizados de manera conjunta realizando descenso de gradiente sobre el límite inferior de evidencia (ELBO). El ELBO es un límite inferior sobre la probabilidad marginal de los datos, también llamado límite inferior variacional. Los gradientes estocásticos necesarios para la optimización son obtenidos mediante una reparametrización.

Modificaciones posteriores se han realizado sobre el modelo inicial tanto para aumentar la expresividad del modelo generativo como para hacer más flexible el modelo de inferencia del posterior. Las primeras destinadas a aumentar la variabilidad de los datos generados para que incluyan toda la diversidad ya codificada en las variable latentes aprendidas por el modelo, las segundas destinadas a aumentar la flexibilidad del modelo de inferencia para hacer que el límite inferior de evidencia más próximo a la probabilidad marginal de los datos y, por tanto, aumentar la representatividad de variables latentes. Otro tipo de propuestas están destinadas a limitar la correlación entre los elementos del vector de variables latentes o incluso a usar una combinación de variables latentes discretas y continuas.

Los VAEs se mantienen como una arquitectura de referencia para la derivación de modelos de inferencia de variables latentes asociadas a conjuntos de variables observadas continuas y discretas y para la síntesis de nuevas instancias pertenecientes a distribuciones de probabilidad próximas a la original.

# Referencias

[1] David H Ackley, Geoffrey E Hinton, and Terrence J Sejnowski. A learning algorithm for boltzmann machines. *Cognitive science*, 9(1):147–169, 1985.

[2] Geoffrey E Hinton and Richard Zemel. Autoencoders, minimum description length and helmholtz free energy. *Advances in neural information processing systems*, 6, 1993.

[3] Umberto Michelucci. An introduction to autoencoders. *arXiv preprint arXiv:2201.03898*, 2022.

[4] Alireza Makhzani and Brendan Frey. K-sparse autoencoders. *arXiv preprint arXiv:1312.5663*, 2013.

[5] Andrew Ng et al. Sparse autoencoder. *CS294A Lecture notes*, 72(2011):1–19, 2011.

[6] Pascal Vincent, Hugo Larochelle, Isabelle Lajoie, Yoshua Bengio, Pierre-Antoine Manzagol, and Léon Bottou. Stacked denoising autoencoders: Learning useful representations in a deep network with a local denoising criterion. *Journal of machine learning research*, 11(12), 2010.

[7] Salah Rifai, Grégoire Mesnil, Pascal Vincent, Xavier Muller, Yoshua Bengio, Yann Dauphin, and Xavier Glorot. Higher order contractive auto-encoder. In *Joint European conference on machine learning and knowledge discovery in databases*, pages 645–660. Springer, 2011.

[8] Charles F Van Loan and G Golub. Matrix computations (johns hopkins studies in mathematical sciences). *Matrix Computations*, 1996.

[9] Elad Plaut. From principal subspaces to principal components with linear autoencoders. *arXiv preprint arXiv:1804.10253*, 2018.

[10] Francisco J Pulgar, Francisco Charte, Antonio J Rivera, and María J Del Jesus. Aeknn: An autoencoder knn-based classifier with built-in dimensionality reduction. *arXiv preprint arXiv:1802.08465*, 2018.

[11] Jinghui Chen, Saket Sathe, Charu Aggarwal, and Deepak Turaga. Outlier detection with autoencoder ensembles. In *Proceedings of the 2017 SIAM international conference on data mining*, pages 90–98. SIAM, 2017.

[12] Chong Zhou and Randy C Paffenroth. Anomaly detection with robust deep autoencoders. In *Proceedings of the 23rd ACM SIGKDD international conference on knowledge discovery and data mining*, pages 665–674, 2017.

[13] Guansong Pang, Chunhua Shen, Longbing Cao, and Anton Van Den Hengel. Deep learning for anomaly detection: A review. *ACM Computing Surveys (CSUR)*, 54(2):1–38, 2021.

[14] Pascal Vincent, Hugo Larochelle, Yoshua Bengio, and Pierre-Antoine Manzagol. Extracting and composing robust features with denoising autoencoders. In *Proceedings of the 25th international conference on Machine learning*, pages 1096–1103, 2008.

[15] Lovedeep Gondara. Medical image denoising using convolutional denoising autoencoders. In *2016 IEEE 16th international conference on data mining workshops (ICDMW)*, pages 241–246. IEEE, 2016.

[16] Michel Broniatowski. Minimum divergence estimators, maximum likelihood and exponential families. *Statistics & Probability Letters*, 93:27–33, 2014.

[17] Benyamin Ghojogh, Ali Ghodsi, Fakhri Karray, and Mark Crowley. Factor analysis, probabilistic principal component analysis, variational inference, and variational autoencoder: Tutorial and survey. *arXiv preprint arXiv:2101.00734*, 2021.

[18] Samuel Gershman and Noah Goodman. Amortized inference in probabilistic reasoning. In *Proceedings of the annual meeting of the cognitive science society*, volume 36, 2014.

[19] Nicholas Metropolis and Stanislaw Ulam. The monte carlo method. *Journal of the American statistical association*, 44(247):335–341, 1949.

[20] Diederik P Kingma and Max Welling. Auto-encoding variational bayes. *arXiv preprint arXiv:1312.6114*, 2013.

[21] Joram Soch, Thomas J. Faulkenberry, Kenneth Petrykowski, and Carsten Allefeld. *The Book of Statistical Proofs*. Zenodo, December 2020. doi:10.5281/zenodo.4305950. URL https://doi.org/10.5281/zenodo.4305950.

[22] Durk P Kingma, Tim Salimans, Rafal Jozefowicz, Xi Chen, Ilya Sutskever, and Max Welling. Improved variational inference with inverse autoregressive flow. *Advances in neural information processing systems*, 29, 2016.

[23] Samuel R Bowman, Luke Vilnis, Oriol Vinyals, Andrew M Dai, Rafal Jozefowicz, and Samy Bengio. Generating sentences from a continuous space. *arXiv preprint arXiv:1511.06349*, 2015.

[24] Anders Boesen Lindbo Larsen, Søren Kaae Sønderby, Hugo Larochelle, and Ole Winther. Autoencoding beyond pixels using a learned similarity metric. In *International conference on machine learning*, pages 1558–1566. PMLR, 2016.

[25] Tim Salimans. A structured variational auto-encoder for learning deep hierarchies of sparse features. *arXiv preprint arXiv:1602.08734*, 2016.

[26] Lars Maaløe, Casper Kaae Sønderby, Søren Kaae Sønderby, and Ole Winther. Auxiliary deep generative models. In *International conference on machine learning*, pages 1445–1453. PMLR, 2016.

[27] Kihyuk Sohn, Honglak Lee, and Xinchen Yan. Learning structured output representation using deep conditional generative models. *Advances in neural information processing systems*, 28, 2015.

[28] Ilya Tolstikhin, Olivier Bousquet, Sylvain Gelly, and Bernhard Schoelkopf. Wasserstein auto-encoders. *arXiv preprint arXiv:1711.01558*, 2017.

[29] Soheil Kolouri, Phillip E Pope, Charles E Martin, and Gustavo K Rohde. Sliced-wasserstein autoencoder: An embarrassingly simple generative model. *arXiv preprint arXiv:1804.01947*, 2018.

[30] Irina Higgins, Loic Matthey, Arka Pal, Christopher Burgess, Xavier Glorot, Matthew Botvinick, Shakir Mohamed, and Alexander Lerchner. beta-vae: Learning basic visual concepts with a constrained variational framework. 2016.

[31] Christopher P Burgess, Irina Higgins, Arka Pal, Loic Matthey, Nick Watters, Guillaume Desjardins, and Alexander Lerchner. Understanding disentangling in beta-vae. *arXiv preprint arXiv:1804.03599*, 2018.

[32] Ricky TQ Chen, Xuechen Li, Roger B Grosse, and David K Duvenaud. Isolating sources of disentanglement in variational autoencoders. *Advances in neural information processing systems*, 31, 2018.

[33] Yuri Burda, Roger Grosse, and Ruslan Salakhutdinov. Importance weighted autoencoders. *arXiv preprint arXiv:1509.00519*, 2015.

[34] Tom Rainforth, Adam Kosiorek, Tuan Anh Le, Chris Maddison, Maximilian Igl, Frank Wood, and Yee Whye Teh. Tighter variational bounds are not necessarily better. In *International Conference on Machine Learning*, pages 4277–4285. PMLR, 2018.

[35] Xianxu Hou, Linlin Shen, Ke Sun, and Guoping Qiu. Deep feature consistent variational autoencoder. In *2017 IEEE winter conference on applications of computer vision (WACV)*, pages 1133–1141. IEEE, 2017.

[36] Jake Snell, Karl Ridgeway, Renjie Liao, Brett D Roads, Michael C Mozer, and Richard S Zemel. Learning to generate images with perceptual similarity metrics. In *2017 IEEE International Conference on Image Processing (ICIP)*, pages 4277–4281. IEEE, 2017.

[37] Eric Jang, Shixiang Gu, and Ben Poole. Categorical reparameterization with gumbel-softmax. *arXiv preprint arXiv:1611.01144*, 2016.

[38] Emilien Dupont. Learning disentangled joint continuous and discrete representations. *Advances in Neural Information Processing Systems*, 31, 2018.

[39] Shengjia Zhao, Jiaming Song, and Stefano Ermon. Infovae: Information maximizing variational autoencoders. *arXiv preprint arXiv:1706.02262*, 2017.

[40] Pengfei Chen, Guangyong Chen, and Shengyu Zhang. Log hyperbolic cosine loss improves variational auto-encoder. 2018.

[41] Abhishek Kumar, Prasanna Sattigeri, and Avinash Balakrishnan. Variational inference of disentangled latent concepts from unlabeled observations. *arXiv preprint arXiv:1711.00848*, 2017.

[42] Aaron Van Den Oord, Oriol Vinyals, et al. Neural discrete representation learning. *Advances in neural information processing systems*, 30, 2017.

[43] Danilo Rezende and Shakir Mohamed. Variational inference with normalizing flows. In *International conference on machine learning*, pages 1530–1538. PMLR, 2015.

[44] Ivan Kobyzev, Simon JD Prince, and Marcus A Brubaker. Normalizing flows: An introduction and review of current methods. *IEEE transactions on pattern analysis and machine intelligence*, 43(11):3964–3979, 2020.

[45] A.K Subramanian. Pytorch-vae. `https://github.com/AntixK/PyTorch-VAE`, 2020.